# Choose qualified instructor for university based on rule-based weighted expert system

**Sana Karimian[1],***

[1]*Jahroom Branch, Islamic Azad University, Jahrom, Fars, Iran*
*\*Corresponding author*

**Abstract:** Near the entire university faculty directors must select some qualified professors for respected courses in each academic semester. In this sense, factors such as teaching experience, academic training, competition, etc. are considered. This work is usually done by experts, such as faculty directors, which is time consuming. Up to now, several semi-automatic systems have been proposed to assist heads. In this article, a fully automatic rule-based expert system is developed. The proposed expert system consists of three main stages. First, the knowledge of human experts is entered and designed as a decision tree. In the second step, an expert system is designed based on the provided rules of the generated decision tree. In the third step, an algorithm is proposed to weight the results of the tree based on the quality of the experts. To improve the performance of the expert system, a majority voting algorithm is developed as a post-process step to select the qualified trainer who satisfies the most expert decision tree for each course. The quality of the proposed expert system is evaluated using real data from Iranian universities. The calculated accuracy rate is 85.55, demonstrating the robustness and accuracy of the proposed system. The proposed system has little computational complexity compared to related efficient works. Also, simple implementation and transparent box are other features of the proposed system.

**Keywords:** Expert systems, Rule based systems, Career recommended system, Qualified instructor choosing

## 1. Introduction

Select qualified instructor for related courses is one of the main tasks which are responded for head of each university faculty. Many different factors are considered by heads to select best instructor between volunteers for a specific course. These are may be: teaching experience, education evidences, ability to communicate well with students, relevant certificates, resume, C.V. and many more. The number of courses which is taught in each semester is too much, so selecting best collection of the instructors isn't an easy problem. For example, 20 to 40 different courses may be taught in computer engineering field at average in each semester. If the number of volunteers is considered "V", instructor collecting may be a hard problem at the exponential order (V20 to V40). In addition, this should be done periodically each semester. An example of instructor selecting problem is as follows:

Suppose three volunteers who are candidate to teach artificial intelligence (AI) course:

• The first candidate has a bachelor's degree in software engineering, a master's degree and a doctorate in artificial intelligence, but he has no teaching experience.

•The second candidate has a bachelor's and master's degree in software engineering and a doctorate in computer vision. He has been teaching electronic topics for 15 years.

• The last candidate has a bachelor's degree in software engineering and a M.Sc. in artificial intelligence, he has taught AI for 3 years. He is currently a doctoral student.

Consider a general feature vector to use, for example, each candidate candidate, which has 4 dimensions that show the certificates and the duration of their teaching (Eq. 1).

$$F_i = \{B.Sc. (Nominal), M.Sc. (Nominal), PhD (Nominal), Education\ experience (Numeric)\} \quad (1)$$

In this respect, three different vectors can be extracted for all candidates in this problem, which are notified by F1, F2 and F3 as follows:

$$F1=\{Software,\ Artificial\ Intelligence,\ Artificial\ Intelligence,\ 0\}$$

$$F2=\{Software,\ Software,\ computer\ vision,\ 15\}$$

$$F3=\{Software,\ Artificial\ Intelligence,\ Null,\ 3\}$$

Given the features extracted above, choosing the best instructor among them is a complex task for the head of the faculty. There are many tips and rules that need to be considered by the faculty head to choose the best ones. Some of the major challenges are as follows:

What features are more important?

Is PhD certificate necessary for teaching AI?

Is it good to choose a volunteer for teaching AI, who has experience just in teaching electronic courses?

Are these features enough to identify a candidate?

This problem is done now by expert humans which is time consuming and low accuracy. Expert systems can be used widely in biochemical [21, 22], medical applications [23, 24], image processing [25, 26], artificial intelligence [27], Healthcare immunology [28], microbiology [31], etc. To satisfy this problems, an intelligent system can be used which may be work based on different intelligent analysis tools such as expert systems, evolutionary algorithms, machine learning, human computer interaction (HCI), statistical pattern recognition algorithms and decision support systems (DSS). In this regard, our research shows that the proprietary algorithm has not yet been proposed to assist faculty heads. An expert system is an approach to artificial intelligence that, based on certain facts and rules, has the ability to replicate the capacity of human thought and reasoning.

The expert system is one of the main approaches to artificial intelligence. Therefore, it is widely used in various applications such as diseases diagnosis [1,2, 17], biology [3], educational [4], recommended systems [5], chemical [6, 15] and so on. The first well known expert system is invented in medical diagnosis titled MYCIN [7] at early 70's. MYCIN was invented at Stanford university to recognize bacteria causing diseases. Gati et al., [8] proposed a method for making better career decisions (MBCD). Proposed MBCD [8], recommend a ranked list of occupations to clients, by considering the client's satisfaction with the chosen occupation. The main part of the proposed MBCD [8] is dialogue processing method which is used to recommend related aspects in terms of client facts. High complexity is one of the main disadvantages of [8] because of using dynamic dialogue instead of static pre-defined rules. Iwayemi et al., proposed a rule based career advisor system. In [9], myers-briggs type indicator (MBTI) is used for facts building based on expert comments. Next, related rules are provided based on MBIT. Finally, a top-down approach is used to inference results. Thakar and Nagori [10], analysis the expert systems for career selection. The use of expert system in career selection offers better efficiency and performance of decision. The analysis is done in [10], based on some innovative factors such as final users, development plan, tools used, knowledge base, inference engine, etc. finally a guideline list is proposed to develop career recommended systems.

In [4], a fully automatic expert system is designed to help head of universities by us. Our results were shown that expert system can be an efficient tool to help faculty heads [4]. In order to improve its quality, our new researches are shown that human experts are not same in terms of experience, course specialty, taught repertoire and etc. In a recent work [12], Okab is proposed an expert system to improve the quality of auditor's performance. The proposed approach in [12], is not absolutely same with our problem, but there is some common points such as weighting human quality and modeling expert knowledge. In this paper, a fully automated expert system is designed to help faculty administrator's to select qualified instructors for each course from among volunteers using a combination of expert weighting techniques and tree classification. Our proposed system consists of three main stages. In the first stage, some decision-made trees are extracted based on facts that are extracted from the expertise of the heads. The representation of knowledge in the designed system is the rule. Therefore, in the second stage, a number of rule sets are extracted from each decision tree produced in the previous stage. Finally, to select the best instructor for each course, a candidate is selected who meets the majority.

## 2. Materials and methods

Almost all expert systems are made up of two parts: the knowledge base (KB) and the inference engine (IE). The first one, knowledge base describes how the expert human knowledge is archived. The second one is IE, which is an engine to conclude results for an input user using KB. These are described as follows. Our designed expert system which helps faculty's head is based on rules which is resulting based on some predefined and extracted rules from cases. Therefore, in the following, the design presented for each of the above two parts will be discussed.

### 2.1. Designed knowledge base

In this problem, heads have to choose the best instructor for each course individually. In this regard, KB must be archived for each course. To do this correctly, we first designed a questionnaire that includes the key factors for each course's lecture. In the next step, the questionnaire is suggested to some experts who are in charge of the facultys in the universities. According to him, all experts will be answered questions. For example, a questionnaire designed for a computer engineering team, completed by a human expert, is shown in Table 1. Note that Table 1 is only completed for example.

In each row, the names of the computer science courses are shown. For example, artificial intelligence (AI), introduction to database (DB), network security (NS), algorithm design (AD), computer networks (CN) are included. Each column shows a specific question that must be answered by a specialist about the related course.

**Table 1. An example of designed questionnaire to teach each lesson in relation of requirements**

| Question / Course | Requirement Certificates | | |
|---|---|---|---|
| | Bs.c. | MS.c. | PhD |
| DB | Software | Software | Software |
| NS | Hardware | Computer Structure | Computer structure |
| AI | Software | Artificial Intelligence | Artificial intelligence |
| CN | Hardware | Computer Structure | Computer structure |
| AD | Software | Algorithm designing or Artificial intelligence | Artificial intelligence |

| Question / Course | Required Taught courses | | | | | Minimum teaching experience |
|---|---|---|---|---|---|---|
| | DB | NS | AI | CN | AD | |
| DB | ■ | | | | | 3 |
| NS | | ■ | | ■ | | 4 |
| AI | | | ■ | | ■ | 5 |
| CN | | | | ■ | | 4 |
| AD | | | | | ■ | 3 |

The answers to the above questionnaire should be in a default values, whether numeric or nominal. For instance, the answer to the question #3 can be just one of the defaults possible fields of PhD computer science such as Artificial Intelligence, Computer Structure or Software. In a same way, three fields are provided for answer to the M.Sc. and PhD questions such as software engineering (SE), Artificial intelligence (AI), algorithm designing (AD) and Computer structure (CS). The questions and related answer values are as follows:

**Table 2. Description and related default answers to the questionnaire**

| Symbol | Description |
|---|---|
| B.Sc. | What should be the B.Sc. field of the instructor's? |
| M.Sc. | What should be the M.Sc. field of instructor's? |
| PhD | What should be the PhD field of instructor's? |
| Teaching experience | How many semesters are required to teach related course? |
| Required Taught courses | What courses should be taught by instructor in previous |
| **Symbol** | **Default response values** |
| B.Sc. | Software / Hardware |
| M.Sc. | Artificial Intelligence/ Computer Structure / Software /Algorithm Designing |
| PhD | Artificial Intelligence/ Computer Structure / Software |
| Teaching experience | Real number |
| Required Taught courses | The Listed Courses (first column of table 1) |

The questionnaires should be completed by experts. Next, a decision tree is designed for each questionnaire. The expert's knowledge is designed and archived using this process. In order to make decision tree, there are many approaches. In this paper, the ID-3 technique is used which have low computational complexity in comparison some other related techniques [13].

The ID-3 makes the tree top-down, in each level, the feature is chosen based on information gain between feature and it's label. In this paper, the decision tree, which is made based on the opinions of the $i_{th}$ human expert, is shown with the notification "$DT_i$". The main contribution of this paper is to provide a rule-based expert system. Therefore, each one of the produced decision trees contains a set of rules. Eventually, by concatenation the produced decision trees to each other, the knowledge base (KB) of the proposed expert system is created as follows:

$$KB = [|DT_1|, |DT_2|, ... , |DT_n|] \quad (2)$$

In figure 1, a decision tree is made using ID-3 technique, for proposed questionnaire that is shown in table 1. As it shown in figure 1, the PhD is chosen as root based on information gain ratio which is more related to the labels than other features.

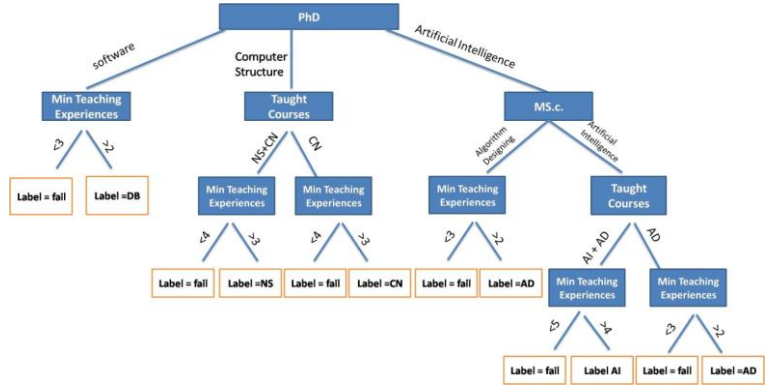

**Fig. 1.** Decision Tree is made based on table 1 results using ID-3 technique

### 2.2. Inference engine
In the previous section, KB is made as combination of some decision trees. Each DT is a full set of direct and inherited rules. In order to design the recommended inference engine (IE), the rules should be extracted from DT's. Extracting rules is done based on proposed algorithm [16]. Each leaf shows posterior of the rule and path from root to converge leaf is show the antecedent of the rule. By extracting related rules, the KB is performed as follows:.
$KB =$

$$[\{R_1^1, R_2^1, \ldots, R_{m1}^1\}, \{R_1^2, R_2^2, \ldots, R_{m2}^2\}, \ldots, \{R_1^n, R_2^n, \ldots, R_{m3}^n\}] \quad (3)$$

Where, $R_i^j$ shows the i$^{th}$ rule that is extracted from j$^{th}$ decision tree. Also, $m_j$ shows the total number of possible extracted rules from j$^{th}$ decision tree. Each one of the rules within the database may assume one or more antecedents. It is very impossible that all of the antecedents satisfied. In many cases, volunteers satisfy only some of the antecedents (not all of them). For each volunteer, the posterior (suggested course) of the rule is selected that has maximum number of satisfied antecedents. For example, suppose a person is candidate to teach in computer science. The facts about the volunteer are shown in the vector format ($F_{query}$) as follows. The rule set that is extracted of respected DT is shown by Rule-Set (DT).

$\mathbf{F_{query}}$ = <Software, Artificial intelligence, Artificial intelligence, {AI, DB, AD}, 4 >

**Table 3.** Volunteer's scores ($F_{query}$) based on completed questionnaire of the table 1

| Rule | Antecedent | Consequent | Firing score for $F_{query}$ |
|---|---|---|---|
| 1 | B.Sc.=software; M.Sc.=AI or AD; PhD=AI; Taught = AD; Experience≥3 | AD | 5 |
| 2 | B.Sc.=software; M.Sc.=software; PhD=software; Taught = DB; Experience≥3 | DB | 3 |
| 3 | B.Sc.=hardware; M.Sc.= structure; PhD= structure; Taught = NS+CN; Experience≥4 | NS | 1 |
| 4 | B.Sc.=software; M.Sc.= AI; PhD= AI; Taught = AI+AD; Experience ≥5 | AI | 4 |
| 5 | B.Sc.=hardware; M.Sc.= structure; PhD= structure; Taught = CN; Experience≥4 | CN | 1 |

As can be seen in the table 3, rule #1 fired higher than some others with score=5. So, the proposed expert systems recommend the volunteer as a good instructor for course AD. It means volunteer can lecture AD better some others. The system recommendation process is show in Eq. (4).
*System Recommend Course =*

$$\{i \mid Max [DT (i)] \quad for\ i=1, 2, \ldots, T_C \} \quad (4)$$

Where, $T_C$ shows the total number of courses in the semester. One of the unique events that happen in the acquisition of knowledge from an expert is providing more than one default answer to the questions.
For example, the expert may consider two factors as answer such as, "Artificial intelligence or computer structure" to the question: What should be the M.Sc. field of instructor's?
To handle this situation, two individual rules can be built with same posterior. Finally, to make decision, the maximum satisfied antecedent between the related rules is considered as DT (Course).

### 2.3. Majority voting
In the above section, the recommendation system was designed based on the decision of one expert. But the aim of this paper is to propose a multi-expert system to increase the final accuracy. In such systems, the decisions of different experts must be combined. In this paper, in order to aggregate the decisions of the experts, an additional voting is conducted between the proposed courses and each course proposed by more experts is selected as the final recommendation.

**2.4. Expert weighting**
In order to improve the quality of designed expert system, an algorithm is proposed in this section which weights experts. Our new researches are shown that human experts are not same in terms of experience, taught courses, etc. In this respect, considering their suggestion with a same weight is not efficient. Suppose, two different experts, first one have taught AI since 20 years before and the second one have taught AI for 2 years and DB for 18 years. Absolutely, if the first one suggests AI for a candidate, his suggestion is more accurate and dependable than second one. Weighting the experts may solve this problem. It can be done in two ways; first way is related to the programmer who can weight expert suggestions based on his desired. This way is not accurate and fact able because of being human dependable. Different functions can be used to weight the suggestion of the experts. In this paper, we consider the related teaching experience as the main factor in weighting process. There are two issues to consider when designing a weighting function:
a.    Determine the threshold for weighing comments. For example, if an expert's teaching experience is less than 5 years, assign the least weight to it.
b.    Use the Gaussian function to weight, because decisions made by experts with age are usually less reliable due to lack of updating.
The proposed expert weighting function is shown in the Eq. 5. Also, it is shown by a simple diagram in figure2.

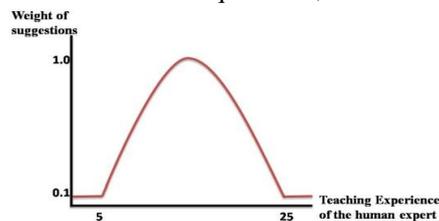

**Fig. 2.** Plot of expert suggestion weighting function

In this respect, the system suggestion after using expert weighting technique can be derived by using the following equation:

$$\textit{System Suggestion} = \{ \ i \ | \ Max \ (DT \ (i) * W(i) \ ) \ for \ i=1,2,\ldots,T_C \ \} \qquad (5)$$

Where, W(i) shows the weight of expert about course "i". Next, the majority voting can be done between the expert suggestions, to achieve final result. The block diagram of designed rule-based weighted expert system is as follows:

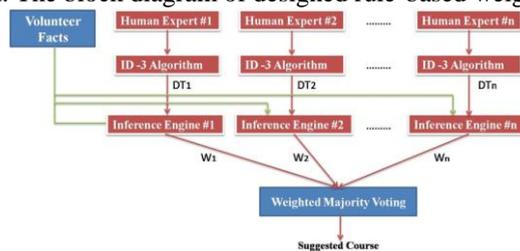

**Fig. 3.** Full flowchart of our proposed rule-based weighted expert system

## 3. RESULTS AND DISCUSSION
The target of this paper is to propose a rule-based expert system to help university faculty heads. In order to evaluate the performance quality, a real train dataset is needed with really data about courses, instructors and their background. A database was collected based on actual information extracted from four different facultys of a same university. The facultys examined in this experiment include the computer science, materials engineering, civil engineering and information technology. On average, 30 samples with 5 features were collected for each faculty. So the total database has 120 samples.
Each sample of the database represents one of the instructors in that faculty. Each one of the features also represents a fact related to that instructor, which is completed based on the questions in the Table 2. The label of each sample also indicates a course title, which the instructor is taught with a very high quality and approved by the faculty heads. Of course, the answer areas are different in each faculty. For example, the question, "What is your B.Sc.?" is same for all of the facultys, but the possible default answer areas are different.
For examples, the answer should be "software / hardware" for computer science. It can be one of two possible statement "electronic" or "communication" in faculty of electronic engineering. Each dataset sample has a label which shows a course name. The faculty heads guaranteed that each sample of the dataset is lectured the related course very well. Notice, dataset samples shows real instructors.

In this respect to evaluate the performance of the proposed system, five different human experts are employed to answer questionnaire. Next, a unique decision-tree is built for each one. In this respect, a train set is made. Finally, the proposed system is runned based on all dataset samples individually. The performance of the proposed system is evaluated based on Eq.6.

$$Acc = N_r / N_{total} \quad (6)$$

Where, $N_r$ shows the number of instances for which the system's suggested lesson is same to the course that is currently being taught successfully. Also, $N_{total}$ shows the total number of database instances that is considered 30 instructors in this experiment. Developing an expert system to help faculty heads is a very recent issue, and the number of articles that have provided an efficient approach to this issue is very small. According to our researches, just Fekri-Ershad et al., [4] presented a method for choosing a instructor based on weightless decision making with the technique of dimensional independence. For comparison, the method [14] has been implemented and the results are presented in the table below.

**Table 4. Performance evaluation of the proposed system for four different faculties**

| Approach / Faculty | [4] | [14] | Proposed approach |
|---|---|---|---|
| **Computer Science** | 83.33 | 82.17 | **86.66** |
| **Material Engineering** | 63.33 | **77.19** | 76.66 |
| **Civil Engineering** | 90.00 | 83.68 | **93.33** |
| **Information Technology** | 75.50 | 76.69 | **81.33** |
| **Average** | 78.04 | 79.93 | **84.49** |

## 4. CONCLUSION

In this paper a full automatic system is designed to help head of university faculties to done their routine tasks with lower cost and higher accuracy. Selecting best instructor and assign better course to them based on their qualities is the main task for heads which is considered in this paper as the main challenge. In this respect, a rule based expert system is developed in this paper. In order to design the proposed system, decision tree algorithm is performed, which makes the system more resistant to possible minor changes between input features. The proposed system is also designed based on the rules, which makes it modular and flexible. The proposed system is designed based on the decision of the majority of experts. This feature allows the user to run some expert systems in parallel and eventually integrate their results to provide more accurate response. The results showed that the reliability and efficiency of the proposed method to offer the best course to the candidates. Low complexity, high accuracy, handling all type of questions as numerical and nominal are the main advantages of our proposed method.